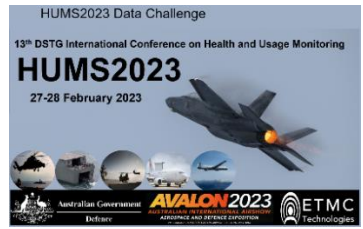

# HUMS2023 Data Challenge Result Submission

**Team Name**: Deakin-MLDS

**Team Members**: Dhiraj Neupane, Lakpa Dorje Tamang, Ngoc Dung Huynh, Mohamed Reda Bouadjenek and Sunil Aryal

**Institutions**: School of IT, Deakin University, Waurn Ponds, VIC, Australia

**Publishable:** Yes

## 1. Summary of Findings

We implemented a simple method for early detection in this research. The implemented methods are plotting the given mat files and analyzing scalogram images generated by performing Continuous Wavelet Transform (CWT) on the samples. Also, finding the mean, standard deviation (STD), and peak-to-peak (P2P) values from each signal also helped detect faulty signs. We have implemented the autoregressive integrated moving average (ARIMA) method to track the progression.

In summary, the earliest signal distortion was seen on the file *Day022_Hunting_SSA_20211209_124241.mat* for sensors Ip-1 and RR-4. For sensor RF-2, the earliest signal fault was detected in *Day022_Hunting_SSA_20211209_141330.mat*. The files *Day027_Hunting_SSA_20220118_111018.mat* and *Day027_Hunting_SSA_20220118_111317.mat* show clear faulty patterns at all four channels. All the observations state that the fault signals were first seen on Day 22, file *Day022_Hunting_SSA_20211209_124241.mat,* in channels 1 and 4. For three out of four sensors' data, except for RL-3, the fault was detected on Day 22, which was not seen further until Day 27. On Day 27, files Day027_Hunting_SSA_20220118_111018.mat and Day027_Hunting_SSA_20220118_111317.mat, the crack was observed clearly. The progression curve, drawn after finding five subsequent values of P2P after Day 27 using ARIMA, shows the abrupt increase in the values after Day 27.

*Table 1    Summary of Analysis Results*

| # | Detection & Trending | Data file name/number | Comments |
|---|---|---|---|
| 1 | Consistent detection on at least one signal channel; i.e. the fault indicators remain consistently above the threshold. | Day022_Hunting_SSA_20211209_141330.mat (RF-2) | 1. Signal Plot [Fig. 4(a)]<br>2. P2p, std and mean analysis (Table 2)<br>3. CWT [Fig. 4(b)] |
| 2 | Confirmed detection on at least two signal channels; i.e. the fault indicators remain consistently above the threshold. | Day022_Hunting_SSA_20211209_124241.mat (IP-1 and RR-4) | 1. Signal Plot [Fig. 2(a) and Fig. 3(a)]<br>2. P2p, std and mean analysis (Table 2)<br>3. CWT [Fig. 2(b) and 3(a)] |
| 3 | Clear multi-channel indication of the characteristic fault features; i.e. faulty planet gear meshing with both the ring and sun gears. | Day027_Hunting_SSA_20220118_111018.mat Day027_Hunting_SSA_20220118_111317.mat (All Channels) | 1. Signal Plot [Fig. 5(a), and Fig. 6(a)]<br>2. P2p, std and mean analysis (Table 2) |

| | | | 3. CWT [Fig 5(b) and Fig. 6 (b)] |
|---|---|---|---|
| 4 | Confirmed trend of fault progression; i.e. a consistent increasing trend started from which file number/name. | Day022_Hunting_SSA_20211209_124241.mat | Fig. 7 and Table 2 |
| 5 | Confirmed trend of accelerated fault progression; i.e. a consistent exponential increasing trend started from which file number/name | Day027_Hunting_SSA_20220118_111018.mat | Fig. 7 and Table 2 |

## 2. Analysis Methods

The methodology implemented in this research is straightforward. The provided datasets were first plotted, and the signal plot was analyzed. The mean, STD and P2P values were calculated from each signal for all four channels and analyzed. The CWT was performed on the samples, and the scalogram images were generated. The generated scalogram images were used to analyze the time-frequency variation of the plotted signals and verify the faults. The signals plot, P2P values, and scalogram images were analyzed to predict the earliest detection. The ARIMA algorithm was implemented to predict five future P2P values. The analysis method is described in detail in section 6.

## 3. Illustrating Figures

This section presents the illustrating figures, which validate our research.

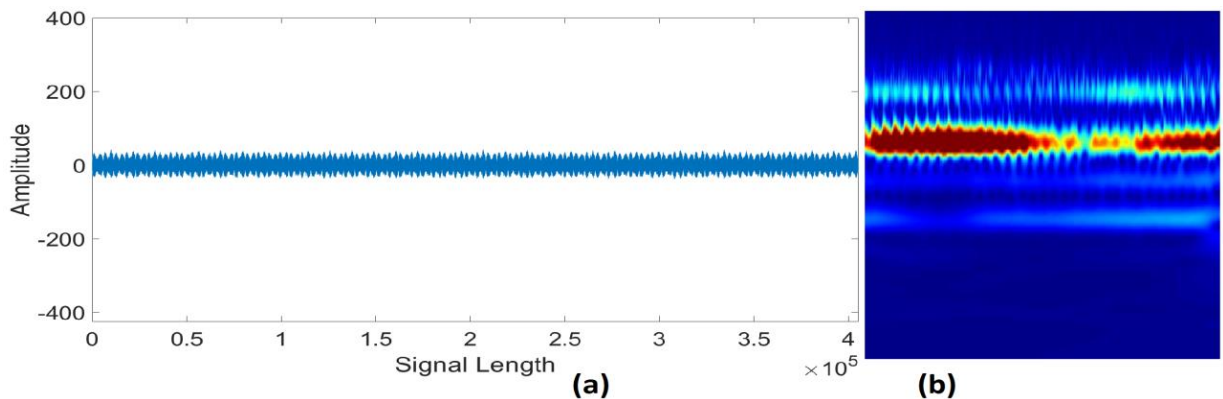

*Fig. 1 Day021_Hunting_SSA_20211208_104755 [Sensor IP-1]: (a) Signal Plot and (b) Scalogram of a sample signal*

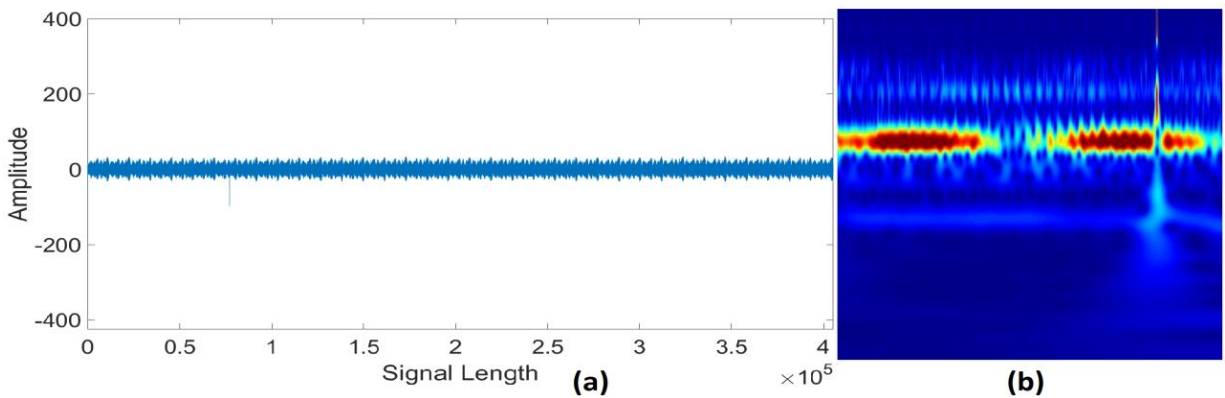

*Fig. 2 Day022_Hunting_SSA_20211209_124241 [Sensor IP-1]: (a) Signal Plot and (b) Scalogram of a sample signal*

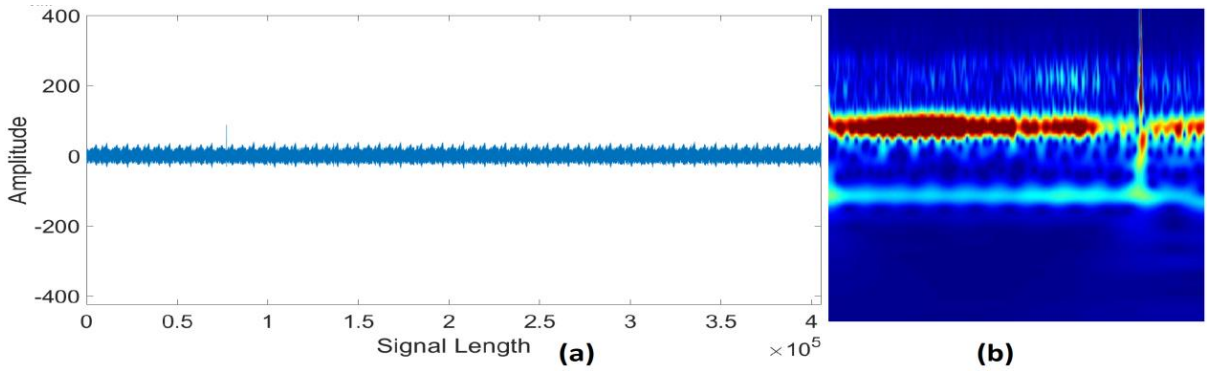

*Fig. 3 Day022_Hunting_SSA_20211209_124241 [Sensor RR-4]: (a) Signal Plot and (b) Scalogram of a sample signal*

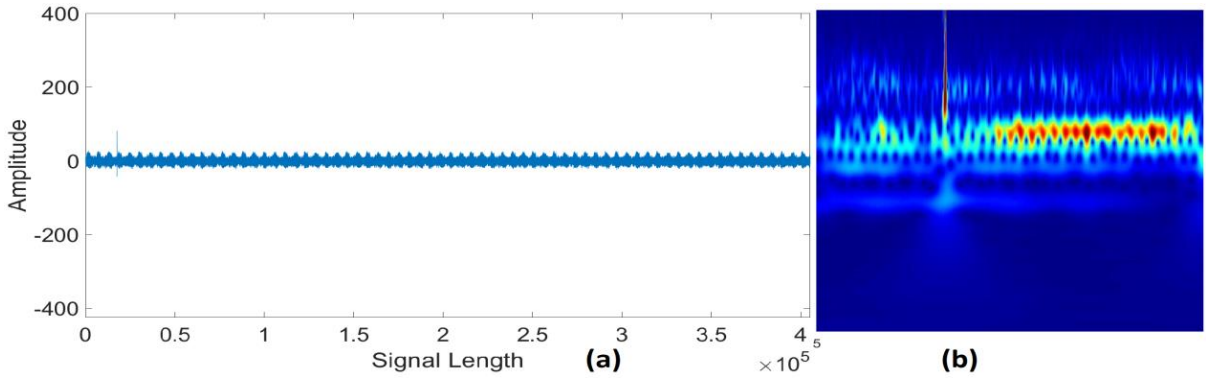

*Fig. 4 Day022_Hunting_SSA_20211209_141330.mat [Sensor RF-2]: (a) Signal Plot and (b) Scalogram of a sample signal*

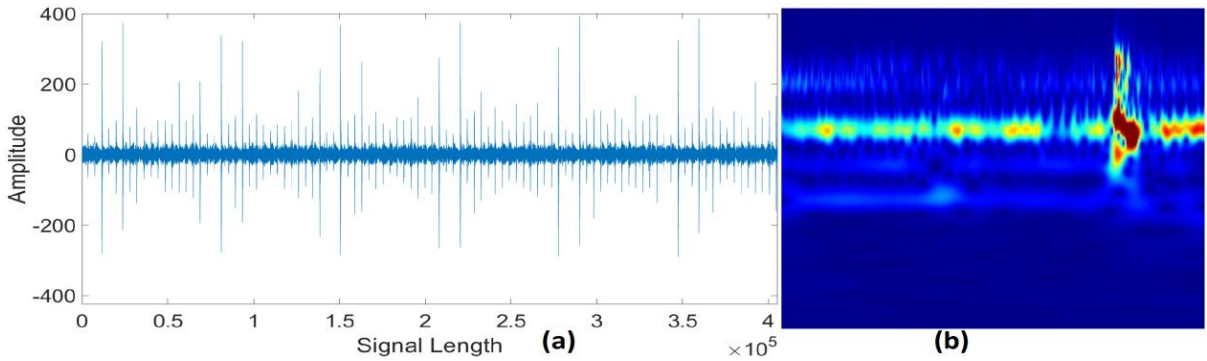

*Fig. 5 Day027_Hunting_SSA_20220118_111018 [Sensor IP-1]: (a) Signal Plot and (b) Scalogram of a sample signal*

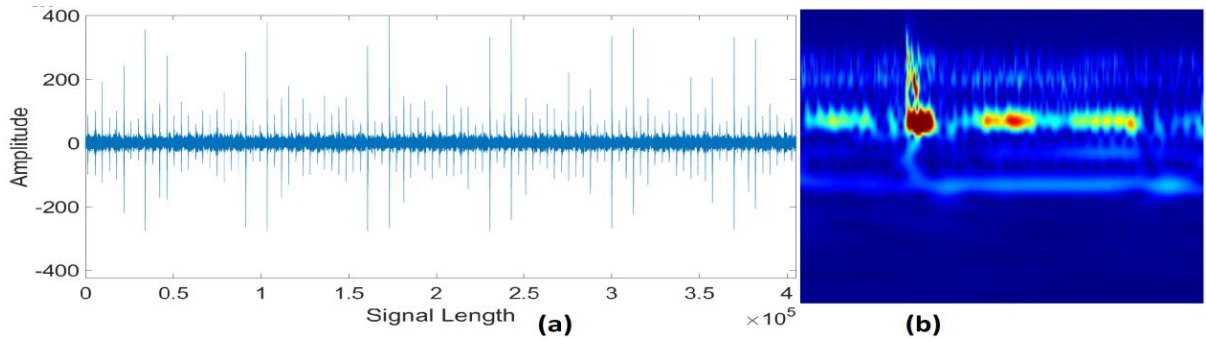

*Fig. 6 Day027_Hunting_SSA_20220118_111317 [Sensor IP-1]: (a) Signal Plot and (b) Scalogram of a sample signal*

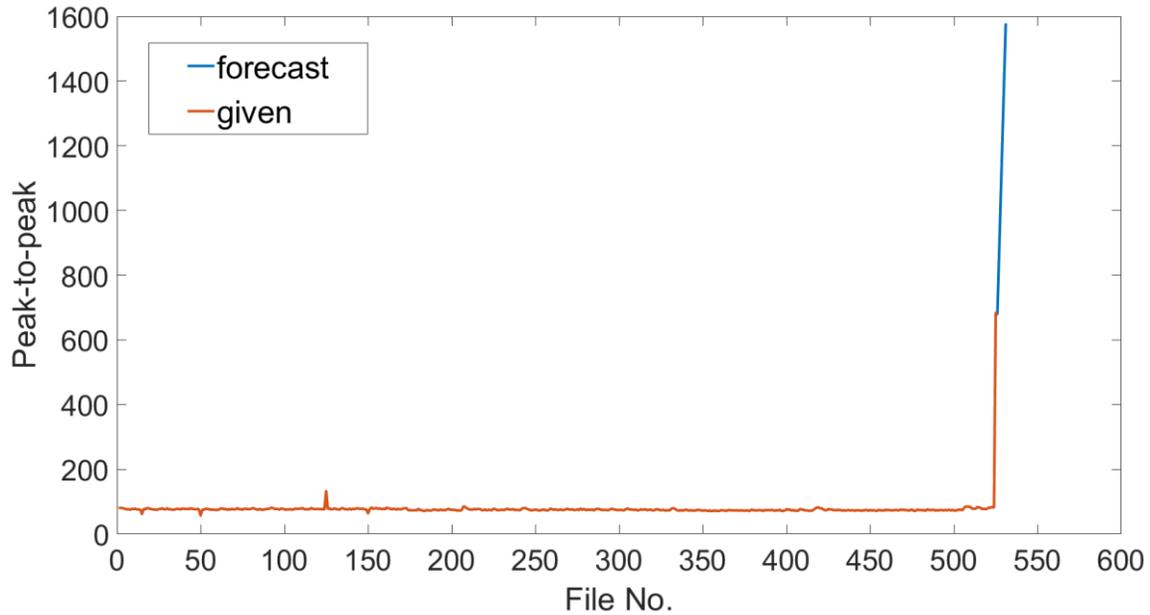

*Fig. 7 Plot of the given and forecasted P2P values for Sensor IP-1*

## 4. Characteristic Fault Signatures of Early Detection

5. Table 2 Top 3 signals of the respective channels with the highest peak-to-peak value

| File Name | Sensor | P2P Value | Remarks | |
|---|---|---|---|---|
| Day021_Hunting_SSA_20211208_104755.mat | IP-1 | 80.8093 | P2P value of the first file of Day 21 (IP-1) | |
| Day022_Hunting_SSA_20211209_124241.mat | IP-1 | 132.9419 | 3rd highest value in IP-1 | Earliest Detection in IP-1 |
| Day027_Hunting_SSA_20220118_111018.mat | IP-1 | 683.2706 | Highest value in IP-1 | |
| Day027_Hunting_SSA_20220118_111317.mat | IP-1 | 679.5497 | 2nd highest value in IP-1 | |
| Day021_Hunting_SSA_20211208_104755.mat | RF-2 | 53.6380 | P2P value of the first file of Day 21 (RF-2) | |
| Day022_Hunting_SSA_20211209_141330.ma | RF-2 | 123.9822 | 3rd highest value in RF-2 | Earliest Detection in RF-2 |
| Day027_Hunting_SSA_20220118_111018.mat | RF-2 | 579.2604 | 2nd highest value in RF-2 | |
| Day027_Hunting_SSA_20220118_111317.mat | RF-2 | 593.4216 | Highest value in RF-2 | |
| Day021_Hunting_SSA_20211208_104755.mat | RL-3 | 81.292285 | P2P value of the first file of Day 21 (RL-3) | |
| Day027_Hunting_SSA_20220118_110720.mat | RL-3 | 107.3236 | 3rd highest value in RL-3 | |
| Day027_Hunting_SSA_20220118_111018.mat | RL-3 | 803.8178 | 2nd highest value in RL-3 | |
| Day027_Hunting_SSA_20220118_111317.mat | RL-3 | 818.7046 | Highest value in RL-3 | |
| Day021_Hunting_SSA_20211208_104755.mat | RR-4 | 93.0710 | P2P value of the first file of Day 21 (RR-4) | |
| Day022_Hunting_SSA_20211209_124241.mat | RR-4 | 125.224648 | 3rd highest value in RR-4 | Earliest Detection in RR-4 |
| Day027_Hunting_SSA_20220118_111018.mat | RR-4 | 609.9475 | 2nd highest value in RR-4 | |
| Day027_Hunting_SSA_20220118_111317.mat | RR-4 | 631.8017 | Highest value in RR-4 | |

The earliest fault detection was observed on the file *Day022_Hunting_SSA_20211209_124241.mat* in the sensors IP-1 and RR-4. For sensor RF-2, the earliest fault in signals was detected in *Day022_Hunting_SSA_20211209_141330.mat*. The files *Day027_Hunting_SSA_20220118_111018.mat* and *Day027_Hunting_SSA_20220118_111317.mat* show clear faulty patterns at all four channels. All the observations state that the fault signals were first seen on Day 22. In three out of four sensors' data, all channels except for sensor RL-3, the fault was detected on Day 22, which was not seen further until Day 27. Files *Day027_Hunting_SSA_20220118_111018.mat* and *Day027_Hunting_SSA_20220118_111317.mat* show a clear crack pattern.

Figure 1 in section 3 shows the signal plot and scalogram image of Day021_Hunting_SSA_20211208_104755 from sensor IP-1. The Signal pattern is observed to be normal, and the corresponding scalogram image also validates the signal's absence of anomaly. In contrast, the signal from Day 22, i.e., file *Day022_Hunting_SSA_20211209_124241.mat* from sensor IP-1, shows some distortion and the presence of faults in figure 2. The amplitude value in the plot, the scalogram image shown in figure 2, and the P2P value from Table 2 validate the abnormality in the pattern. Thus, *Day022_Hunting_SSA_20211209_124241.mat* file is detected as the earliest faulty signal for sensor IP-1. Moving on, the plot and scalogram of Day 27 files *Day027_Hunting_SSA_20220118_111018.mat* and *Day027_Hunting_SSA_20220118_111317.mat* in figures 5 and 6 show the signal pattern, which is very different from the rest. Also, the P2P values of these files in Table 2 are incredibly high, validating our point.

## 5. Fault Progression Trending Curve

The fault progression tracking was problematic because of the signal's characteristics. Except for plotting each signal (4 sensors times 526 files), we also plotted the 526 signals from each sensor as one signal and analyzed the pattern. To our surprise, the signal's pattern was different for a few files only, which were identified as the earliest detected faulty signals in the earlier sections. The signal followed the same pattern until the last two files from Day 27, where the signal was found entirely defective. The plot images can be found in our GitHub repository.

The track progression method implemented in this research is ARIMA. The fault progression curve for sensor IP-1 is shown in figure 7, which is the plot of given values and the following five predicted values. The X-axis is the number of files (the given file number is 526, predicted is 5; thus total=531), and Y-axis shows the P2P value. The images showing the progression tracking of other sensors are uploaded to our GitHub repository.

## 6. Description of Analysis Methods

### Description of fault detection method

As mentioned earlier, signal plotting, scalogram image generation using continuous wavelet transform, and analyzing mean, STD and P2P values are the techniques used in finding the earliest detection in this research. The provided datasets were first plotted, and the signal plot was analyzed. Also, CWT was performed on the samples generated from splitting the signal further to obtain the scalogram images. The scalogram is a plot of the absolute value of the CWT signal as a time and frequency function.

For the generation of scalogram images, an analytical Morlet (amor) wavelet of the CWT family was used. Various studies [1, 2] have shown that the machinery's vibration signals of the machinery like bearings and gears consist of periodic impulses. The impact signals of the faulty machinery signals match well with that of the Morlet wavelet. That is why the Morlet wavelets indicate machinery defects and healthy bearing characteristics in a well-defined manner, and as a result, they are widely used in this field. The analysis method can be divided into two sub-parts:

### Pre-processing:

As a data pre-processing, each file with a length of 405405 data points is split into 98 samples, each of length 4096. Thus, each file is further divided into 98 smaller signals. For instance, the initial 4096 data points make the first sample signal, then another 4096 data points, i.e., the data points from 4097 to 8192, make the second sample signal, and the process continues until 98 samples are made. Out of the total length of each file, the first 401408 data points are used in generating the sample signals. The last 3997 data points are neglected as the number does not fulfil the sample length chosen.

### Scalogram generation:

As mentioned, scalogram images were generated from each pre-processed sample, with 4096 data points, using an analytical Morlet wavelet of the CWT family. The *cwtfilterbank* function of MATLAB is used to obtain the two-dimensional time-frequency images of size 500×500. The number of voices per octave, $v$, is 12. In CWT, some base is fixed, which is a fractional power of two; for example, $2^{1/v}$, where $v$ is a parameter known as 'voices per octave'. It is a term commonly used to designate the number of wavelet filters per octave. In any audio or musical signal, the number of octaves determines the span of frequencies being analyzed, while the number of voices per octave determines the number of samples (scales) across this span. Different scales are obtained by raising this base scale to positive integer powers. The reason $v$ is referred to as the number of voices per octave is that increasing the scale by an octave (a doubling) finer the discretization of the scale parameter. However, this also increases the computation required because the CWT must be computed for every scale [3].

### Description of fault trending method

The algorithm used to track fault progression in this research is ARIMA, a time-series model that identifies hidden patterns in time-series data and makes forecasts. The auto-regression sub-model in ARIMA uses past values for forecasting. The integrated sub-model performs differencing to remove any non-stationary in the time series. The moving average sub-model uses past errors to make a prediction. The progression curve obtained by plotting the given and predicted P2P values for sensor IP-1 is shown in figure 7.

In a study [4], the authors claim that using mean, STD, and P2P values gives the best performance while detecting the faults in the machinery data. Thus, we have used the timestamp, STD, mean, and P2P values as the features for training the ARIMA model. We used all the 526 features as training data and predicted the following five P2P values.

## 7. Supplement Information

As the supplement information, we have uploaded the Jupyter notebook file containing plots of all the signals (526 files times four sensors), the excel file containing the mean, STD, and P2P values of all the signals, the scalogram images of the signals mentioned in Table 2, the plot of given and forecasted P2P values for all sensors, and the full-length plot (526 files in a single frame) of all sensors in our GitHub repository. The link to the repository is https://github.com/dhirajneupane/HUMS2023.

## References

[1] J. Guo, X. Liu, S. Li, and Z. Wang, "Bearing Intelligent Fault Diagnosis Based on Wavelet Transform and Convolutional Neural Network," *Shock and Vibration*, vol. 2020, p. e6380486, Nov. 2020, doi: 10.1155/2020/6380486.

[2] Y. Yoo and J.G. Baek, "A Novel Image Feature for the Remaining Useful Lifetime Prediction of Bearings Based on Continuous Wavelet Transform and Convolutional Neural Network," *Applied Sciences*, vol. 8, no. 7, p. 1102, Jul. 2018, doi: 10.3390/app8071102.

[3] D. Neupane, Y. Kim, and J. Seok, "Bearing Fault Detection Using Scalogram and Switchable Normalization-Based CNN (SN-CNN)," *IEEE Access*, vol. 9, pp. 88151–88166, 2021, doi: 10.1109/access.2021.3089698.

[4] W. Wang, K. Vos, J. Taylor, C. Jenkins, B. Bala, L. Whitehead and Z. Peng, "Is Deep Learning Superior in Machine Health Monitoring Applications," *12th DST International Conference on Health and Usage Monitoring*, 29 November 2021 -1 December 2021, Melbourne, Australia